%% file: main.tex
\documentclass[runningheads]{llncs}

\usepackage[mobile]{eccv}

\usepackage{eccvabbrv}

\usepackage{graphicx}
\usepackage{tabularx} 
\usepackage{booktabs}

\usepackage[accsupp]{axessibility}  

\usepackage[pagebackref,breaklinks,colorlinks,citecolor=eccvblue]{hyperref}

\usepackage{orcidlink}
\usepackage[accsupp]{axessibility}
\usepackage{graphicx}
\usepackage{amsmath}
\usepackage{amssymb}
\usepackage{bbm}
\usepackage{booktabs}
\usepackage{xfrac}
\usepackage{multirow}
\usepackage{pifont}
\usepackage{colortbl}
\usepackage{setspace}
\usepackage{pgfplots}
\pgfplotsset{compat=1.17} 

\usepackage[noend]{algpseudocode}
\usepackage{algorithm}
\usepackage{listings}
\usepackage{amsfonts}       
\usepackage{nicefrac}       
\usepackage{microtype}      
\usepackage{xcolor}         

\definecolor{codeblue}{rgb}{0.25,0.5,0.5}

\algnewcommand\algorithmicinput{\textbf{Input:}}
\algnewcommand\INPUT{\item[\algorithmicinput]}
\algnewcommand\algorithmicoutput{\textbf{Output:}}
\algnewcommand\OUTPUT{\item[\algorithmicoutput]}

\usepackage{breqn}
\usepackage{caption}
\usepackage{subcaption}
\usepackage{wasysym}

\usepackage{stfloats}
\usepackage{float}
\usepackage{makecell}
\usepackage{wrapfig,lipsum}
\usepackage{adjustbox}
\usepackage{pythonhighlight}

\usepackage[capitalize]{cleveref}
\crefname{section}{Sec.}{Secs.}
\Crefname{section}{Section}{Sections}
\Crefname{table}{Table}{Tables}
\crefname{table}{Tab.}{Tabs.}
\crefname{equation}{Eq.}{Eqs.}

\newcommand{\cmark}{\ding{51}}
\newcommand{\name}{VISAGE}
\newcolumntype{Y}{>{\centering\arraybackslash}X}

\begin{document}

\title{VISAGE: Video Instance Segmentation with Appearance-Guided Enhancement} 

\titlerunning{Video Instance Segmentation with Appearance-Guided Enhancement}

\author{Hanjung Kim\inst{1} \and
Jaehyun Kang\inst{1} \and
Miran Heo\inst{1} \and
Sukjun Hwang\inst{2} \and \\
Seoung Wug Oh\inst{3} \and
Seon Joo Kim\inst{1}}

\authorrunning{H. Kim et al.}

\institute{Yonsei University \and
Carnegie Mellon University \and
Adobe Research}

\maketitle

\input{sec/0_abstract}
\input{sec/1_intro}
\input{sec/2_related_works}
\input{sec/3_method}
\input{sec/4_experiments}
\input{sec/5_limitations}
\input{sec/6_conclusion}
\clearpage  

%
%
\bibliographystyle{splncs04}
\bibliography{main}
\end{document}

%% file: sec/0_abstract.tex
\begin{abstract}
In recent years, online Video Instance Segmentation (VIS) methods have shown remarkable advancement with their powerful query-based detectors. 
Utilizing the output queries of the detector at the frame-level, these methods achieve high accuracy on challenging benchmarks.
However, our observations demonstrate that these methods heavily rely on location information, which often causes incorrect associations between objects. 
This paper presents that a key axis of object matching in trackers is appearance information, which becomes greatly instructive under conditions where positional cues are insufficient for distinguishing their identities.
Therefore, we suggest a simple yet powerful extension to object decoders that explicitly extract embeddings from backbone features and drive queries to capture the appearances of objects, which greatly enhances instance association accuracy.
Furthermore, recognizing the limitations of existing benchmarks in fully evaluating appearance awareness, we have constructed a synthetic dataset to rigorously validate our method.
By effectively resolving the over-reliance on location information, we achieve state-of-the-art results on YouTube-VIS 2019/2021 and Occluded VIS (OVIS).
Code is available at \href{https://github.com/KimHanjung/VISAGE}{https://github.com/KimHanjung/VISAGE}.

\keywords{video instance segmentation, object appearance}

\end{abstract}

%% file: sec/1_intro.tex
\section{Introduction}
\label{sec:intro}

Video Instance Segmentation (VIS) is a challenging task that requires classification, segmentation, and tracking of distinct instances throughout a video sequence~\cite{MaskTrackRCNN}.
Current studies in VIS can be primarily categorized into two approaches: online and offline, based on whether a video is processed in a per-frame or per-clip manner. 
Recently, the advancement of frame-level object detectors has resulted in online methods becoming increasingly dominant in the VIS field.

As detectors directly impact the accuracy in the video domain, recent online models are primarily built using the powerful query-based detectors~\cite{Deformable-DETR, Mask2Former}.
Spatially decoding image information, the detectors are designed to represent object-wise information using the queries.
Therefore, the online VIS methods reuse these queries from the detectors and have achieved substantial improvements in multiple challenging benchmarks~\cite{OVIS-Dataset, vis2022} by mostly adopting either propagation~\cite{GenVIS, TCOVIS} or matching~\cite{MinVIS, IDOL, CTVIS} strategies.
However, tracking under complex scenarios such as \textit{shot changes} or \textit{trajectory intersections} (\cref{fig:teaser}) remains imperfect, resulting in the degradation of the overall accuracy.

\input{fig/teaser}

Examining these failure cases, we observe that \textit{object-wise information} of the queries is significantly imbalanced: heavy reliance on positional cues, and less reflection on appearances.
As demonstrated in \cref{fig:teaser}, previous query-based VIS methods~\cite{GenVIS, CTVIS, MinVIS, IDOL, DVIS}, tend to maintain the spatial order of previous predictions in their current predictions.
To support this argument, we conduct additional analyses by horizontally flipping images to generate two-frame pseudo videos.
The existing models manifest association errors despite the distinct exterior patterns of objects, as shown at the top of \cref{fig:pseudo_flip}, which highlight the dependence on object locations.
As there exist multiple scenarios that cannot be fully handled with the imbalanced information, such a phenomenon necessitates the models to take object appearances into consideration.

We introduce \textbf{VISAGE} (\textbf{V}ideo \textbf{I}nstance \textbf{S}egmentation with \textbf{A}ppearance-\textbf{G}uided \textbf{E}nhancement), a method that leverages appearance cues as a crucial indicator for distinguishing instances.
In our approach, we introduce a streamlined branch that employs mask pooling to generate \textit{appearance} queries from the predicted mask of \textit{object} queries.
This enables each appearance query to capture the visual features of its corresponding object, providing a more comprehensive representation for improved tracking accuracy.
To refine query discrimination, we integrate a contrastive loss~\cite{ContrastiveNIPS, SimCLR, QDTrack_TPAMI}, which enhances the model's ability to distinguish between instances across different frames.

\input{fig/pseudo_flip}

Additionally, we introduce a streamlined tracker designed to minimize reliance on heuristic procedures to the greatest extent feasible.
Previous methods~\cite{IDOL, CTVIS} employ multiple refinement steps for removing redundant mask proposals and adopt handcrafted threshold values for accurate tracklet construction.
Through such processing, only selectively chosen queries are incorporated into the matching process. 
Although these complex tracker configurations enhance tracking accuracy, they also create a dependency on numerous hyperparameters, each of which can be tailored heuristically to specific datasets.
To alleviate such dependence, our method streamlines the tracker and dramatically reduces the number of required hyperparameters, such as threshold values for initializing and deleting tracklets, and non-maximum suppression (NMS), among others.
Nonetheless, a lack of temporal information still exists, inherent in query-matching online methods, as they are only aware of adjacent frames.
We address this limitation by using a simple memory bank to facilitate temporal awareness.

Despite the simplicity of our approach, \name{} has many desirable properties.
Our method introduces a new paradigm in query-based VIS by emphasizing the crucial role of appearance information for object association. 
Enhanced by appearance-based guidance, it demonstrates superior performance in complex tracking scenarios, outperforming previous methods that often misidentify objects due to an excessive dependence on spatial information as shown in \cref{fig:teaser}(a). 
It successfully leverages appearance information, as illustrated at the bottom of \cref{fig:pseudo_flip}, and has been validated on our proposed large-scale pseudo dataset, outperforming established methods~\cite{GenVIS, CTVIS} by a large margin.
Furthermore, with its simplified tracker that effectively utilizes the past history of both object and appearance queries, our method demonstrates competitive performance across all benchmark datasets.
Notably, \name{} achieves state-of-the-art performance on three standard benchmarks: YouTubeVIS-19/21~\cite{MaskTrackRCNN, vis2021} and OVIS~\cite{OVIS-Dataset}.

%% file: fig/teaser.tex


\begin{figure}[t]
\begin{center}
\includegraphics[width=\linewidth]{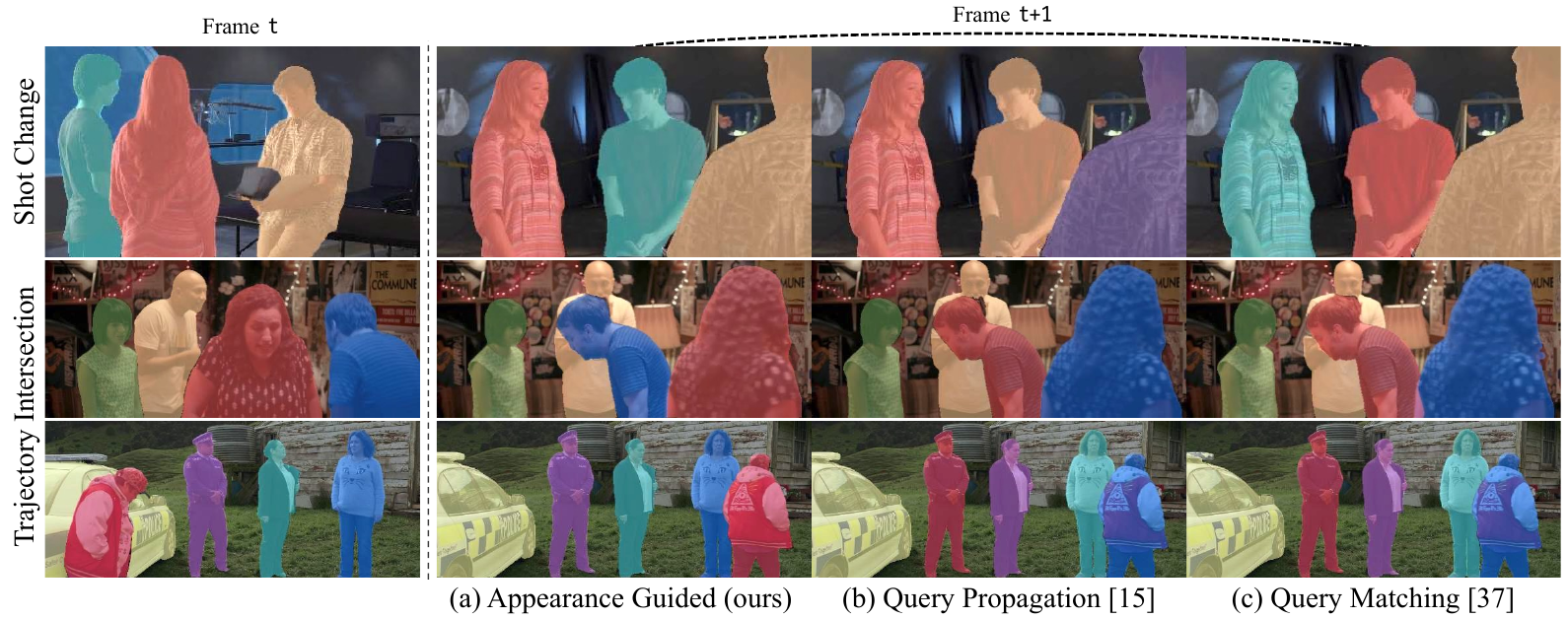}
\end{center}
\vspace{-6mm}
\caption{
\textbf{Qualitative results across challenging scenarios.} 
Predicted results using query-propagation~\cite{GenVIS, DVIS}, query-matching~\cite{MinVIS,CTVIS,IDOL}, and our appearance-guided methods. 
The first row illustrates a shot change across consecutive frames, a scenario where previous methods fail to maintain consistent tracking. 
The second and third rows demonstrate trajectory intersections, leading to id-switching with previous methods. 
Unlike previous methods, our method successfully tracks objects without switching or losses.
Best viewed in color.
}
\vspace{-7mm}
\label{fig:teaser}
\end{figure}

%% file: fig/pseudo_flip.tex

\begin{figure}[ht]
\begin{center}
\includegraphics[width=0.7\linewidth]{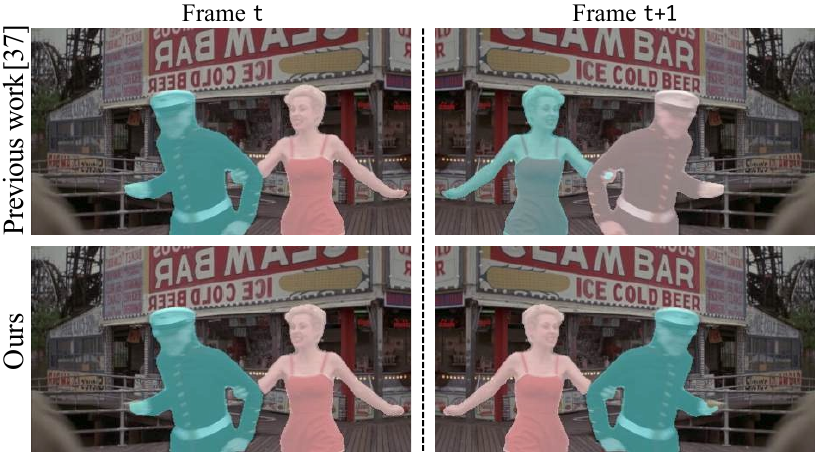}
\end{center}
\vspace{-5mm}
\caption{
\textbf{Proof of concept demonstrated with a flipped image.}
Previous methods~\cite{MinVIS, CTVIS, DVIS, GenVIS, IDOL} struggle with instance matching in flipped images, showing a dependency on location. 
Our method, \name{}, addresses this by emphasizing appearance, enabling accurate instance matching even with image flipping.
}
\vspace{-4mm}
\label{fig:pseudo_flip}
\end{figure}

%% file: sec/2_related_works.tex
\section{Related Works}

\subsection{Online Video Instance Segmentation}
Contrary to traditional online approaches~\cite{MaskTrackRCNN, CrossVIS, VISOLO, PCAN}, modern online methods utilize query-based detectors~\cite{DETR, Deformable-DETR, Mask2Former} with an emphasis on query association strategies. 
Query-based online methods can be divided into two main strategies: query-matching and query-propagation.

\textbf{Query-matching} approaches dynamically construct tracklets in an online fashion using query-based detectors ~\cite{DETR, Deformable-DETR, Mask2Former}, which yield predictions for each video frame individually. 
MinVIS~\cite{MinVIS} implements this concept by exclusively training of the query-based detector, subsequently deploying it on video frames independently during the inference phase to conduct tracking via bipartite matching of corresponding queries.
However, its supervision is restricted to the frame-level, which can introduce ambiguity by not accounting for the object's continuity in the video sequence.
To mitigate this, some studies~\cite{IDOL, CTVIS} have incorporated contrastive learning~\cite{ContrastiveNIPS, SimCLR, QDTrack_TPAMI} to refine instance embeddings.
Moreover, these methods utilize a memory bank at the inference stage, which allows for the processing of multiple frames, thereby enhancing the temporal information captured for each object.
Specifically, CTVIS~\cite{CTVIS} further improves discriminative capability by using the memory bank during the training stage.
However, these previous approaches still fall short in discriminative ability due to their insufficient use of appearance information. 
Our method, \name{}, introduces a novel strategy that utilizes the appearance cue to significantly enhance discriminative ability, leading to more robust tracking.

\textbf{Query-propagation} methods track objects in video sequences by utilizing output queries from prior frames.
By propagating output queries, these methods track corresponding objects across frames~\cite{CAROQ}. 
Additionally, some enhance tracking accuracy by also propagating proposals alongside the queries~\cite{InsPro}.
Recent developments have enabled some methods to operate in both online and offline modes, processing videos clip-by-clip and frame-by-frame. 
For instance, GenVIS~\cite{GenVIS} employs an offline VIS method as its backbone, wherein learnable instance prototypes aggregate the backbone's outputs through the propagation of the instance prototype. 
This design allows GenVIS to function as either an online or offline method, depending on the length of clip processed by the backbone. 
On the other hand, DVIS~\cite{DVIS} constructs tracklets by propagating frame queries in an online manner. These well-established tracklets are then refined to effectively utilize information from the entire video in an offline manner.
Yet, these methods are often constrained by a local matching strategy that focuses on aligning tracklets with ground truths based solely on the current frame. 
This approach frequently leads to unstable tracking outcomes, as it fails to consider the entire video context, resulting in suboptimal performance.
TCOVIS~\cite{TCOVIS} addresses this issue by shifting from the local matching strategy of GenVIS~\cite{GenVIS} to a global matching approach, thereby achieving more robust tracking across videos.

\subsection{Offline Video Instance Segmentation}
Offline VIS architectures~\cite{StEm-Seg, EfficientVIS} process input videos at the clip-level rather than frame by frame.
VisTR~\cite{VisTR} extends DETR~\cite{DETR} from frame-level to clip-level processing in an end-to-end manner by simultaneously handling multiple frame features within the transformer encoder-decoder. 
However, this approach, which processes multiple frame-level inputs at once, demands extensive computation, making the processing of longer sequences impractical.

To overcome this limitation, IFC~\cite{IFC} introduces a memory token in the transformer encoder and employs fixed-size clip queries in the transformer decoder, enhancing both the model's performance and efficiency. 
With the transformer decoder design of IFC, Mask2Former-VIS~\cite{Mask2Former-VIS} adapts Mask2Former~\cite{Mask2Former} for video-level tasks, resulting in substantial performance gains. 
Meanwhile, SeqFormer~\cite{SeqFormer} redesigns the transformer decoder to process each frame individually, thus improving the model's ability to detect instance movement by precisely capturing location changes. 
VITA~\cite{VITA} introduces a novel strategy  by processing clip queries through cross-attention with frame queries rather than relying on frame features.
This strategy lightens the computational load imposed by dense spatio-temporal backbones, resulting in an efficient architecture capable of managing lengthy videos.
Beyond focusing solely on transformer architecture enhancements, TeViT~\cite{TeViT} introduces a groundbreaking backbone that improves temporal information processing by capitalizing on the strengths of ViT~\cite{ViT}.

%% file: sec/3_method.tex
\section{Method}
We now present \name{}.
Initially, we provide an overview of the query-based detector architecture upon which \name{} is constructed. 
Then, we introduce a novel strategy that improves object association through appearance guidance.
Lastly, we detail our inference pipeline with a simplified structure that significantly enhances performance.

\subsection{Query-based Detector}
Query-based object detectors~\cite{DETR, Deformable-DETR, Mask2Former} can be largely divided into three components: backbone, transformer encoder, and transformer decoder.
The backbone initiates the process by generating low-level image feature maps, encapsulating essential visual information. 
These feature maps are then enhanced through the transformer encoder, which employs self-attention mechanisms to refine the feature representation, as detailed in~\cite{DETR, Deformable-DETR}.
The process concludes in the transformer decoder, where the identified objects are decoded into $N$ learnable queries.
Our approach adopts the well-established query-based detector framework~\cite{Mask2Former}, maintaining its original structure intact.

\subsection{Appearance-Guided Enhancement}
Prior to the advent of query-based tracking approaches, traditional video tracking methods~\cite{MaskTrackRCNN, QDTrack_TPAMI, VPS} extracted instance features from backbone feature maps using operations such as RoIPool~\cite{FastRCNN} and RoIAlign~\cite{MaskRCNN}. 
Following a similar principle, we use average pooling to extract appearance queries from the backbone feature maps, guided by the predicted masks as shown in \cref{fig:main} (a).
These appearance queries are designed to encapsulate appearance-centric features, providing a distinctive complement to the object queries.
Consequently, alongside the object queries already in use, we introduce appearance queries as an additional indicator.
We then transform both types of queries into \textit{appearance embeddings} $\mathbf{e}_a\in\mathbb{R}^{N\times C}$ and \textit{object embeddings} $\mathbf{e}_i\in\mathbb{R}^{N\times C}$, respectively.

\input{fig/main_figure}

Our appearance embeddings enhance the matching process when using only object embeddings leads to incorrect matches, as illustrated in~\cref{fig:main} (c).
Relying solely on the similarity of object embeddings may lead to ambiguity due to their similar positions across subsequent frames.
However, this ambiguity can be resolved by also considering the similarity of appearance embeddings, as their distinct appearances provide additional discriminative information.
With this guidance, we leverage both appearance embeddings $\mathbf{e}_a$ and object embeddings $\mathbf{e}_i$ to identify the optimal match between queries across subsequent frames as shown in \cref{alg:inference} (lines 4-10).

To enhance object association quality, we improve the distinctiveness of both object and appearance embeddings.
We utilize contrastive learning to refine embeddings obtained from two distinct frames, ensuring that embeddings of identical object instances are brought closer together in the embedding space, while those of different instances are separated further apart.
Unlike previous methods utilizing contrastive learning~\cite{IDOL,QDTrack_TPAMI}, our approach treats object and appearance embeddings individually, applying contrastive loss to each respectively.
It allows each type of embedding to be distinctly characterized by its inherent properties, facilitating their mutual synergy in object matching, as illustrated in \cref{fig:main} (c).
Consequently, our model's final loss integrates a weighted sum of the contrastive losses for both types of embeddings with the original query-based detector's loss.

\subsection{Inference with Appearance}
As shown in~\cref{alg:inference}, our tracker employs a simple yet effective matching process that aligns the current object embeddings and appearance embeddings with their respective counterparts from the previous frame. 
We compute the similarity scores for each object-appearance embedding pair using Cosine Similarity, following a method similar to that in \cite{MinVIS}. 
Then, employing the weighted sum of object and appearance similarities, we utilize the Hungarian algorithm \cite{Hungarian} (\texttt{linear_sum_assignment} in~\cref{alg:inference}) to achieve optimal assignment.

\input{alg/inference}

In addition, we incorporate a simple memory bank to compensate for the lack of temporal information, a limitation inherent in online processes, as used in previous methods~\cite{IDOL, CTVIS}.
We stack the states of previous queries from the most recent frames within our memory bank, which has a size of $W$. 
From this bank, we read a memory embedding $m\in\mathbb{R}^{N\times C}$ using the \texttt{read_memory()} function in~\cref{alg:inference}.
Specifically, this function is implemented by temporally weighting the embeddings to put more emphasis on recent queries while utilizing the confidence scores for selective weighting.
For the current memory embedding $m^t$, the calculation of the weighting for each embedding at the previous time step $w\in[1, W]$ can be formally expressed as follows:
\begin{equation}
    \mathbf{m}^t = \sum\limits_{w=1}^{W}\left(\mathbf{e}^{t-w}s^{t-w} \times \frac{W}{w}\right),
\end{equation}
where $s$ denotes the confidence score.
This memory embedding represents the object and appearance, denoted as $m^t_i$ and $m^t_a$ respectively, 
in~\cref{alg:inference}. 

Finally, we sequentially associate the predictions ($\mathbf{p}$ in~\cref{alg:inference}) from each frame using the obtained assignment.
By employing a simple inference pipeline coupled with an efficient memory bank, we introduce an effective approach.

%% file: fig/main_figure.tex
\begin{figure*}[t]
\begin{center}
   \includegraphics[width=\linewidth, bb=0 0 850 332]{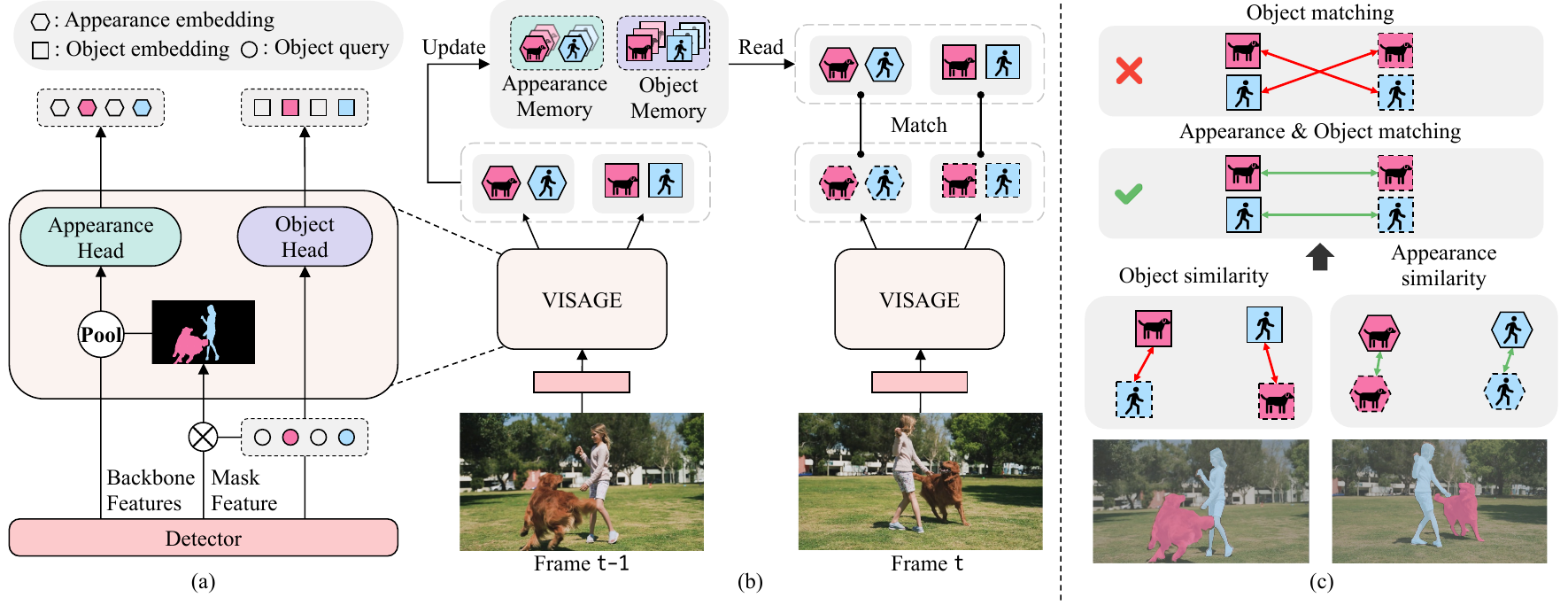}
\end{center}
\vspace{-6mm}
\caption{ \textbf{Overview of ~\name{}.}
(a) The proposed ~\name{}'s architecture which generate object embedding and appearance embedding.
(b) Overall inference pipeline of ~\name{}: At time step $t-1$, the memory bank is updated with both the appearance embedding and the object embedding. 
Then, at time step $t$, the memory embedding is read from the memory bank and used for matching.
(c). Details of the matching process: In that scenario, using only object embeddings leads to incorrect matching. 
However, when guided by the appearance embedding, the matching process can be corrected. Best viewed in color.
}
\vspace{-5mm}
\label{fig:main}
\end{figure*}

%% file: alg/inference.tex
\begin{wrapfigure}{r}{0.57\textwidth}
\vspace{-8mm}
\begin{minipage}{0.55\textwidth}
\vspace{-7mm}
\begin{algorithm}[H]
\caption{Inference pipeline of \name{}.}
\label{alg:inference}
\begin{algorithmic}[1]
\INPUT model $\mathcal{F}$, frames $\{x^t\}_{t=1}^{T}$, weight $\alpha$
\OUTPUT predictions $\mathcal{P}$

\State \( \mathcal{P} \gets \{\} \)
\State \( \mathcal{M} \gets \{\} \) 
\State \( \texttt{idx} \gets [0, N-1] \)
\For{$t$ in $[1, T]$\textbf{:}}
        \State \( {\mathbf{p}}^t, \mathbf{e}_{i}^t, \mathbf{e}_{a}^t \gets \mathcal{F}(x^t)\)
        
        \State \(\mathcal{M} \gets \mathcal{M} + (\mathbf{e}_{i}^t[\texttt{idx}], \mathbf{e}_{a}^t[\texttt{idx}]) \) 
        \State \(\mathbf{m}_{i}^t, \mathbf{m}_{a}^t \gets \mathcal{M}\texttt{.read_memory()}\)
        \State \(\mathcal{P} \gets \mathcal{P} + {\mathbf{p}}^t[\texttt{idx}] \)

        
        \State \( \mathbf{s} \gets (1-\alpha)\cdot\texttt{cos_sim}(\mathbf{e}_{i}^t, \mathbf{m}_{i}^t)\)
        \Statex \phantom{\ \ \ \ \ \ \ \ }\( +\ \alpha\cdot\texttt{cos_sim}(\mathbf{e}_{a}^t, \mathbf{m}_{a}^t)\) 
        \State \( \texttt{idx} \gets \texttt{linear_sum_assignment}(\mathbf{s})\)
\EndFor 
\State \textbf{end for}
\State \textbf{return} $\mathcal{P}$
\end{algorithmic}
\end{algorithm}
\end{minipage}
\vspace{-10mm}
\end{wrapfigure}


        
        

%% file: sec/4_experiments.tex
\section{Experiments}

\subsection{Datasets}
We evaluate \name{} on four VIS benchmarks: Youtube-VIS (YTVIS) 2019 / 2021 / 2022 \cite{MaskTrackRCNN, vis2021, vis2022} and Occluded VIS (OVIS) \cite{OVIS-Dataset}.
The YTVIS datasets contain 40 predefined categories in their videos. 
YTVIS 2019 is the first and largest dataset for video instance segmentation.
It includes 2,238 videos for training, 302 for validation, and 343 for the testing.
YTVIS 2021 expands upon the YTVIS 2019 dataset, containing 2,985 videos for training, 421 for validation, and 453 for testing, while refining annotations and modifying some categories.
Additionally, to represent challenging scenarios with longer sequences, YTVIS 2022 is added. 
It adds 71 extra videos to the validation set of YTVIS 2021.

OVIS stands out for its more complex and longer videos, featuring 25 categories and comprising 607 videos for training, 140 for validation, and 154 for testing.
In comparison to YTVIS, it contains a higher number of instances per video, averaging 5.8, and a total of 296k masks. 
Additionally, the average length of videos in OVIS is approximately 12 seconds.

\subsection{Implementation Details}
We adopt the Mask2Former~\cite{Mask2Former} as our query-based detector.
All of our models are initialized with parameters pre-trained on COCO dataset~\cite{COCO} with ResNet-50~\cite{ResNet} backbone.
We also adopt COCO joint training following previous works~\cite{SeqFormer, VITA, GenVIS, CTVIS, TCOVIS}.
Our batch includes 16 videos.
In our experimental setup, we set the weights for the contrastive losses applied to both the appearance and object embeddings at $2.0$.
For losses other than the contrastive loss, we adopt the same loss function and weight specifications as those described in \cite{Mask2Former-VIS}.
The window size of memory bank $W$ is set to 5 and appearance weight $\alpha$ is set to 0.75 during inference.
Finally, our method is trained using 4 NVIDIA A6000 GPUs.

\input{tab/ytvis_2019_2021_2022}

\subsection{Main Results}
We compare our methods with state-of-the-art methods on four VIS benchmarks: YTVIS 2019/2021/2022 and OVIS.
The results are reported in \cref{tab:ytvis2019_2021_2022} and \cref{tab:ovis}.

\subsubsection{Youtube-VIS 2019 \& 2021 \& 2022.}
As shown in \cref{tab:ytvis2019_2021_2022}, we compare our \name{} with previous state-of-the-art methods.
On the YTVIS 2019 benchmark, \name{} performs on par with the highest-performing method cited as CTVIS~\cite{CTVIS}. 
Notably, both \name{} and CTVIS~\cite{CTVIS} surpass previous methods by a large margin. 
On the YTVIS 2021 benchmark, an improved version of YTVIS 2019, \name{} outperforms other existing methods. 
We achieve the highest performance on both the YTVIS 2019 and 2021 datasets by incorporating appearance-guided enhancement. 
Finally, on the YTVIS 2022 benchmark, our \name{} exhibits comparable performance compared to GenVIS\cite{GenVIS} and TCOVIS~\cite{TCOVIS}.

\subsubsection{OVIS.}

In \cref{tab:ovis}, we present a comparison on the OVIS benchmark~\cite{OVIS},  which is characterized by long videos and complex scenarios, including frequent occlusions.
\name{} achieves  performance on par with the previously established state-of-the-art methods such as GenVIS~\cite{GenVIS}, TCOVIS~\cite{TCOVIS}, and CTVIS~\cite{CTVIS}.
Through our proposed properties, which include appearance-guided enhancement and a simplified tracker, \name{} effectively handles such long and complicated videos.
As a result, these advancements enable \name{} to achieve state-of-the-art performance.

\input{tab/ovis}

\subsection{Ablation Studies}
In this section, we provide the ablation studies for our proposed method and discuss its effect.
All ablation experiments are conducted on YTVIS 2019~\cite{MaskTrackRCNN} validation set.

\subsubsection{Appearance feature.}
In~\cref{tab:ablation_featurepool}, we analyze the impact of feature maps which create the appearance query.
When we make the appearance query from the transformer encoder feature maps, there is a degradation in performance.
This indicates that backbone feature maps contain rich visual information compared to transformer encoder feature maps.

\subsubsection{Appearance guidance.}
\cref{tab:ablation_component} shows effectiveness of our appearance-guided enhancement.
We omit appearance information by setting the $\alpha$ in line 9 of \cref{alg:inference} to 0, resulting in a matching process that relies solely on object similarity. 
As indicated by rows 1 and 3 in~\cref{tab:ablation_component}, the absence of appearance information leads to a degradation in performance. 
Further analysis of the effectiveness of appearance guidance is discussed in~\cref{sec:appearance}.

\subsubsection{Memory bank.}
In~\cref{tab:ablation_component}, we demonstrate the necessity of the memory bank for compensating for the lack of temporal information. 
We evaluate performance differences between scenarios with and without the use of the memory bank. 
Rows 1 and 2 in \cref{tab:ablation_component} show decreased performance compared to rows 3 and 4, respectively.
Without a memory bank, our method is constrained to using only the immediately preceding frame for historical context, relying on similarities measured exclusively between consecutive frames. 
However, the introduction of a memory bank expands this capability by leveraging a broader historical perspective.
Furthermore, the adoption of a memory bank significantly boosts the effectiveness of appearance information.
This is because historical appearance information provides a more reliable basis for matching than only considering the appearance from the immediate preceding frame to recognize an identical object.
By implementing a memory bank, our method gains awareness of previous frames, which leads to improved performance.

\input{tab/ablations}

\subsubsection{Tracklet processing.}
Previous query-matching based methods~\cite{IDOL, CTVIS}, build upon the tracker framework from \cite{QDTrack_TPAMI}, incorporating handcrafted thresholding and heuristic post-processing techniques, including Non-Maximum Suppression (NMS).
In contrast, our tracker exclusively utilizes cosine similarity and the Hungarian algorithm for matching, as detailed in \cref{alg:inference}.
To understand the impact of handcrafted designs, we align our inference pipeline with those of methods previously used ~\cite{IDOL, CTVIS}.
The results, as detailed in \cref{tab:ablation_tracker}, reveal that our streamlined approach consistently surpasses the performance of the conventional tracker.
A detailed examination, especially between rows 1 and 2 of \cref{tab:ablation_tracker}, reveals that the absence of heuristic elements, such as NMS, leads to a decline in performance.
However, row 3 reveals that our simplified tracker performs impressively even without heuristic design elements.
This suggests that a simplified tracker is capable of achieving commendable performance without the complexity of heuristic design elements.

\subsection{Analysis of Appearance-Guided Enhancement}

\label{sec:appearance}
\subsubsection{Appearance Weight.}
In our methodology, as detailed in line 9 of \cref{alg:inference}, the hyperparameter $\alpha$ plays a crucial role by representing the weighting of the appearance similarity.
By tuning the value of $\alpha$, we modulate the emphasis placed on the appearance information.
In \cref{tab:ablation_component}, we demonstrate how our appearance guidance significantly enhances performance on the YTVIS 2019 dataset. 
In this section, we further analysis the impact of appearance cue on robust matching by exploring various values of $\alpha$.

\cref{tab:appearance} highlights the critical role of both appearance and location information in tracking. 
We analyze the impact of appearance weight $\alpha$ on both the YTVIS 2019 and OVIS validation sets.
There are two extreme cases to consider: relying solely on location information or exclusively on appearance information. 
We observe a reduced performance on both datasets when the $\alpha$ value is set to 0, with this reduction being more pronounced on the OVIS dataset.
Given OVIS's complex scenarios, such as frequent occlusions, the dataset's intricacies make the contribution of appearance information especially significant.
Conversely, setting $\alpha$ to 1, and thus relying only on appearance information, results in a significant drop in performance.
Without the positional cue, the model only depends on appearance information for matching across frames, leading to potential ambiguities in making object tracklets. 

\input{tab/analysis_appearance}

However, integrating both appearance and location information consistently surpasses these two extreme cases, highlighting the complementary strengths of using both cues in establishing robust tracklets. 
Notably, increasing the value of $\alpha$, and thereby the emphasis on appearance information, correlates with performance enhancements. 

\subsubsection{T-SNE Visualization.}
As illustrated in \cref{fig:tsne}, we analyze the impact of appearance-guided enhancement in \name{} on its association capabilities using the OVIS dataset by visualizing the query embeddings for an entire video, with consistent colors indicating the same instance.
The visualization includes three types of query embeddings: object-only, appearance-only, and a combination of both object and appearance queries. 
While object-only embeddings for the same object exhibit scattered and inconsistent clustering, appearance-only query embeddings demonstrate a high degree of clustering.
The integration of object and appearance query embeddings results in even more distinct and pronounced clustering, as exemplified by the red circle in \cref{fig:tsne}. 
This enhanced clustering clearly indicates that our appearance-guided enhancement leads to better association.

\subsubsection{Pseudo Dataset.}
Traditional datasets do not fully cover the complex scenarios that our approach is designed to address. 
To validate \name{} in a more intuitive manner, we construct a pseudo dataset consisting of synthetic videos.
Instances in pseudo videos are created by compositing objects from the COCO dataset~\cite{COCO} with Copy-and-Paste augmentation~\cite{CopyPaste}. 
The background images for the pseudo dataset are randomly sourced from BG-20k~\cite{BG-20K}.
Additionally, we employ Bezier Curves to simulate the movement of objects.
It includes two types of videos: \textit{track}, where objects move following arbitrary Bezier Curves, and \textit{swap}, where objects' locations are randomly swapped along their trajectories.
Except for the movement of instances, both types of datasets are generated under the same conditions.

\input{fig/tsne_pseudo}

As shown at the top of \cref{fig:pseudo}, instances in the track type pseudo video move along a Bezier Curve. 
Consequently, complex scenarios, such as intersections between instances or movements out of the frame, occur naturally.
On the other hand, the swap type presents a challenge for methods that primarily rely on location information, as this dependence results in incorrect matching. 
This is illustrated at the bottom of \cref{fig:pseudo}, where the positions of instances in the pseudo video are suddenly swapped. 
Such scenarios verify the method's awareness of appearance cues when location information is no longer a reliable indicator.

In ~\cref{tab:pseudo}, we conduct an evaluation of various online VIS methods~\cite{GenVIS, CTVIS} including \name{} on the pseudo dataset using published ResNet-50 backbone weights, which are trained on the YouTube-VIS 2019 dataset.
All of these methods use COCO joint training, ensuring a fair comparison.
The overall trend is similar to that observed on the YTVIS 2019 dataset.
In contrast, \name{} outperforms other methods in swap type videos, mainly due to the challenges posed by the dataset in associating instances using only location information. 
Consequently, other methods demonstrate degraded performance on swap type videos compared to their performance on track types.
Furthermore, the absence of appearance information in the matching process notably impacts \name{}'s performance on swap type videos, although it remains relatively unaffected for track type videos.
It proves that appearance cue serves as an additional indicator, providing robustness in scenarios with swapped object positions. 

\input{tab/pseudo}

\subsection{Qualitative Results}
\cref{fig:qualitative} displays the visualization results of our \name{} across various challenging scenarios. 
In the first row, several cats are positioned closely together and the red-colored cat crosses another cat.
In such cases, relying solely on location information can often result in identity-switching errors. 
In the second row, we present a similar scenario involving different classes: a person and a cow.

\input{fig/qualitative}

In the third row, we showcase another challenging scenario where an object completely disappears and then reappears: a cat disappears and then reappears. 
Our memory bank, which stores the previous queries of the cat, enables us to reestablish the association when the cat reappears. 
The fourth row demonstrates a similar scenario, where our memory bank proves effective: a small sedan is completely occluded by the truck. 
Even though the large truck in the foreground and the small, disappeared car are located closely, their distinct appearances ensure accurate matching.

%% file: tab/ytvis_2019_2021_2022.tex
\begin{table}[t]
\centering
\caption{
Comparisons on the \textbf{YouTube-VIS 2019, 2021 and 2022 validation} sets. 
Methods are denoted as online or offline, indicated by the text color.
We highlight the best performance in \textbf{bold}.
}
\vspace{-1mm}
\resizebox{\textwidth}{!}
{ 
\begin{tabular}{@{}lc|c|ccccc|ccccc|ccccc@{}}
\toprule
\multicolumn{2}{l|}{\multirow{2}{*}{Method}}                        & \multirow{2}{*}{Setting} & \multicolumn{5}{c|}{YouTube-VIS 2019} & \multicolumn{5}{c|}{YouTube-VIS 2021} & \multicolumn{5}{c}{YouTube-VIS 2022}\\
\multicolumn{2}{l|}{}                                               &         & AP        & AP$_{50}$ & AP$_{75}$ & AR$_1$    & AR$_{10}$ & AP        & AP$_{50}$ & AP$_{75}$ & AR$_1$    & AR$_{10}$ & AP        & AP$_{50}$ & AP$_{75}$ & AR$_1$    & AR$_{10}$ \\
    \midrule
    \midrule

     \multicolumn{2}{l|}{IFC~\cite{IFC}}                           & \textcolor{gray}{offline}
                                                                                    & 41.2      & 65.1      & 44.6      & 42.3      & 49.6
                                                                                    & 35.2      & 55.9      & 37.7      & 32.6      & 42.9  
                                                                                    & -         & -         & -         & -         & -    \\
     \multicolumn{2}{l|}{Mask2Former-VIS~\cite{Mask2Former-VIS}}   & \textcolor{gray}{offline}
                                                                                    & 46.4      & 68.0      & 50.0      & -         & -  
                                                                                    & 40.6      & 60.9      & 41.8      & -         & -     
                                                                                    & -         & -         & -         & -         & -    \\
     \multicolumn{2}{l|}{MinVIS~\cite{MinVIS}}                     & online         & 47.4      & 69.0      & 52.1      & 45.7      & 55.7      
                                                                                    & 44.2      & 66.0      & 48.1      & 39.2      & 51.7  
                                                                                    & 23.3      & 47.9      & 19.3      & 20.2      & 28.0  \\
     \multicolumn{2}{l|}{IDOL~\cite{IDOL}}                         & online         & 49.5      & 74.0      & 52.9      & 47.7      & 58.7   
                                                                                    & 43.9      & 68.0      & 49.6      & 38.0      & 50.9  
                                                                                    & -         & -         & -         & -         & -    \\
     \multicolumn{2}{l|}{VITA~\cite{VITA}}                         & \textcolor{gray}{offline}
                                                                                    & 49.8      & 72.6      & 54.5      & 49.4      & 61.0
                                                                                    & 45.7      & 67.4      & 49.5      & 40.9      & 53.6  
                                                                                    & 32.6      & 53.9      & 39.3      & 30.3      & 42.6  \\
     \multicolumn{2}{l|}{GenVIS~\cite{GenVIS}}                     & online         & 50.0      & 71.5      & 54.6      & 49.5      & 59.7
                                                                                    & 47.1      & 67.5      & 51.5      & 41.6      & 54.7  
                                                                                    & 37.5      & \textbf{61.6}      & 41.5      & 32.6      & 42.2  \\
     \multicolumn{2}{l|}{GenVIS~\cite{GenVIS}}                     & \textcolor{gray}{offline}
                                                                                    & 51.3      & 72.0      & 57.8      & 49.5      & 60.0
                                                                                    & 46.3      & 67.0      & 50.2      & 40.6      & 53.2  
                                                                                    & 37.2      & 58.5      & \textbf{42.9}      & 33.2      & 40.4  \\
     \multicolumn{2}{l|}{DVIS~\cite{DVIS}}                         & online         & 51.2      & 73.8      & 57.1      & 47.2      & 59.3
                                                                                    & 46.4      & 68.4      & 49.6      & 39.7      & 53.5  
                                                                                    & -         & -         & -         & -         & -    \\
     \multicolumn{2}{l|}{DVIS~\cite{DVIS}}                         & \textcolor{gray}{offline}
                                                                                    & 52.6      & 76.5      & 58.2      & 47.4      & 60.4
                                                                                    & 47.4      & 71.0      & 51.6      & 39.9      & 55.2  
                                                                                    & -         & -         & -         & -         & -    \\
     \multicolumn{2}{l|}{TCOVIS~\cite{TCOVIS}}                     & online         & 52.3      & 73.5      & 57.6      & 49.8      & 60.2
                                                                                    & 49.5      & 71.2      & 53.8      & 41.3      & 55.9  
                                                                                    & \textbf{38.6}      & 59.4      & 41.6      & 32.8      & \textbf{46.7}  \\
     \multicolumn{2}{l|}{CTVIS~\cite{CTVIS}}                       & online         & \textbf{55.1}      & \textbf{78.2}      & 59.1      & \textbf{51.9}      & \textbf{63.2}
                                                                                    & 50.1      & 73.7& 54.7      & 41.8      & \textbf{59.5}  
                                                                                    & -         & -         & -         & -         & -    \\
     \multicolumn{2}{l|}{\textbf{\name{}}}                         & online         & \textbf{55.1}      & 78.1      & \textbf{60.6}      & 51.0      & 62.3
                                                                                    & \textbf{51.6}      & \textbf{73.8}& \textbf{56.1}      & \textbf{43.6}      & 59.3 
                                                                                    & 37.5      & 60.0      & 37.1      & \textbf{35.2}      & 44.1  \\
    \bottomrule
    \end{tabular}
 } 
\vspace{-5mm}
\label{tab:ytvis2019_2021_2022}
\end{table}

%% file: tab/ovis.tex
\begin{table}[t]
\centering
\caption{
Comparison on the \textbf{OVIS validation} set. 
\textbf{Bold} indicates the best performance. 
We use the online version for GenVIS~\cite{GenVIS} and DVIS~\cite{DVIS}.
}
\vspace{-2mm}
\resizebox{0.5\textwidth}{!}{ 
\begin{tabular}{@{}lc|ccccc@{}}
\toprule
\multicolumn{2}{l|}{\multirow{2}{*}{Method}}                        & \multicolumn{5}{c}{OVIS} \\
\multicolumn{2}{l|}{}                                               & AP        & AP$_{50}$ & AP$_{75}$ & AR$_1$    & AR$_{10}$ \\
    \midrule
    \midrule

     \multicolumn{2}{l|}{MinVIS~\cite{MinVIS}}                     & 25.0      & 45.5      & 24.0      & 13.9      & 29.7  \\
     \multicolumn{2}{l|}{IDOL~\cite{IDOL}}                         & 30.2      & 51.3      & 30.0      & 15.0      & 37.5  \\
     \multicolumn{2}{l|}{GenVIS~\cite{GenVIS}}                     & 35.8      & \textbf{60.8}      & 36.2      & 16.3      
                                                                                    & 39.6  \\
     \multicolumn{2}{l|}{DVIS~\cite{DVIS}}                         & 31.0      & 54.8      & 31.9      & 15.2      & 37.6   \\
     \multicolumn{2}{l|}{TCOVIS~\cite{TCOVIS}}                     & 35.3      & 60.7      & \textbf{36.6}      & 15.7      & 39.5   \\
     \multicolumn{2}{l|}{CTVIS~\cite{CTVIS}}                       & 35.5      & \textbf{60.8}      & 34.9      & 16.1      
                                                                                    & \textbf{41.9} \\
     \multicolumn{2}{l|}{\textbf{\name{}}}                         & \textbf{36.2}      & 60.3      & 35.3      & \textbf{17.0} & 40.3 \\
    \bottomrule
    \end{tabular}}
\vspace{-3mm}
\label{tab:ovis}
\end{table}

%% file: tab/ablations.tex
\begin{table}[t]
  \centering
  \caption{Ablation study of the target feature maps which generate appearance query}
  \vspace{-2mm}
  \begin{tabular}{@{}l|ccccc@{}}
    \toprule
    Feature & AP & AP$_{50}$ & AP$_{75}$ & AR$_1$ & AR$_{10}$ \\
    \midrule
    Transformer Encoder & 51.4 & 72.4 & 56.7 & 49.8 & 60.7 \\
    Backbone & 55.1      & 78.1      & 60.6      & 51.0      & 62.3   \\
    \bottomrule
  \end{tabular}
  \label{tab:ablation_featurepool}
  \vspace{-1mm}
\end{table}

\begin{table}[t]
  \begin{minipage}{.45\textwidth}
    \footnotesize
    \caption{Ablation study on appearance guidance and memory bank utilization, with memory bank window size $W=5$.}
    \vspace{-2mm}
    \begin{tabular}{@{}cc|ccccc@{}}
\toprule
\multirow{2}{*}{\shortstack{App}} &   \multirow{2}{*}{\shortstack{Memory\\Bank}} &   \multicolumn{5}{c}{YouTube-VIS 2019}\\
           &           & AP        & AP$_{50}$ & AP$_{75}$ & AR$_1$    & AR$_{10}$ \\
    \midrule
    \midrule
           &           & 49.9      & 71.4      & 54.7      & 47.0      & 58.7   \\
 \cmark    &           & 50.2      & 72.1      & 54.7      & 47.3      & 60.7   \\
           & \cmark    & 53.4      & 76.8      & 58.7      & 49.8      & 61.2   \\
 \cmark    & \cmark    & 55.1      & 78.1      & 60.6      & 51.0      & 62.3   \\
    \bottomrule
    \end{tabular}
    \vspace{1mm}
    \label{tab:ablation_component}
  \end{minipage}
  \hspace{6mm}
  \begin{minipage}{.45\linewidth}
    \centering
    \footnotesize
    \caption{Ablation study of the tracklet post-processing.}
    \vspace{-2mm}
    \begin{tabular}{@{}cc|ccccc@{}}
\toprule
Tracker            & NMS    & AP        & AP$_{50}$ & AP$_{75}$ & AR$_1$    & AR$_{10}$ \\
    \midrule
    \midrule

\cite{IDOL, CTVIS} &  & 52.7      & 75.3      & 57.6      & 49.3      & 61.6      \\
\cite{IDOL, CTVIS} & \cmark & 54.4      & 77.9      & 59.0      & 49.9      & 62.8      \\
ours               &  & 55.1      & 78.1      & 60.6      & 51.0      & 62.3   \\
    \bottomrule
    \end{tabular}
    \vspace{1mm}
    \label{tab:ablation_tracker}
  \end{minipage}
  \vspace{-5mm}
\end{table}

%% file: tab/analysis_appearance.tex
\begin{table}[t]
  \centering
  \caption{\textbf{Analysis of Appearance Weight $\alpha$.} The plot demonstrates the relationship between different appearance weight $\alpha$ settings and their corresponding AP scores on the YouTube-VIS 2019 and OVIS validation sets.}
  \begin{tabularx}{0.8\textwidth}{Y|YYYYY|YYYYY}
        \toprule
        \multirow{2}{*}{\shortstack{$\alpha$}} &   \multicolumn{5}{c|}{YouTube-VIS 2019} &   \multicolumn{5}{c}{OVIS}\\
                    & AP        & AP$_{50}$ & AP$_{75}$ & AR$_1$    & AR$_{10}$      & AP        & AP$_{50}$ & AP$_{75}$ & AR$_1$    & AR$_{10}$      \\
            \midrule
            \midrule
          0.00      & 53.4      & 76.8      & 58.7      & 49.8      & 61.2      
                    & 32.2      & 55.6      & 30.9      & 16.1      & 36.3      \\
          0.25      & 53.6      & 77.0      & 59.4      & 50.0      & 61.1      
                    & 34.5      & 59.2      & 32.5      & 16.7      & 38.8      \\
          0.50      & 54.5      & 77.2      & 60.3      & 50.9      & 61.7      
                    & 34.8      & 59.8      & 35.3      & 16.2      & 39.4      \\
          \rowcolor{gray!=25}
          0.75      & 55.1      & 78.1      & 60.6      & 51.0      & 62.3      
                    & 36.2      & 60.3      & 35.3      & 17.0      & 40.3      \\
          1.00      & 24.9      & 36.7      & 28.7      & 40.9      & 53.7      
                    & 11.4      & 22.7      & 10.2      & 10.3      & 24.9      \\
        \bottomrule
    \end{tabularx}
    \label{tab:appearance}
    \vspace{-3mm}
\end{table}

%% file: fig/tsne_pseudo.tex
\begin{figure}[t]
  \begin{minipage}{.45\textwidth}
    \centering
    \includegraphics[width=1.0\linewidth]{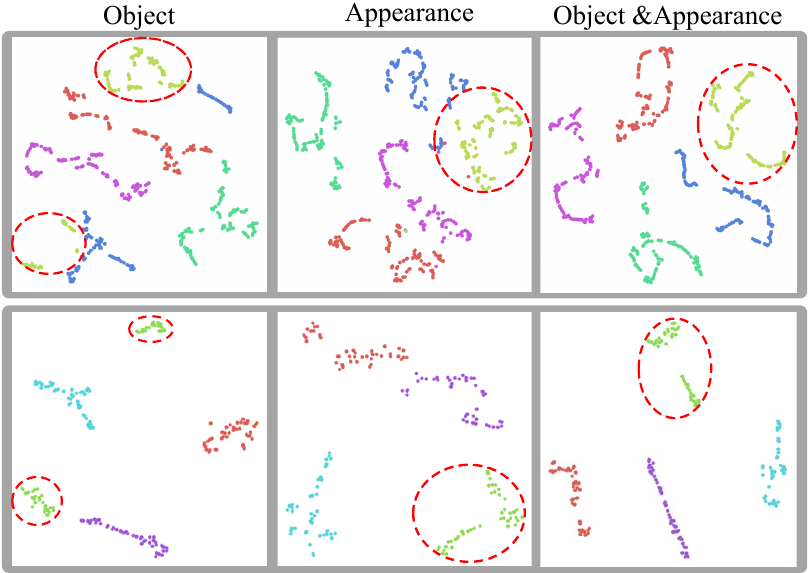}
    \caption{
    \textbf{T-SNE visualization on the OVIS dataset.}
    Each row representing three different videos. 
    Each column corresponds to the type of query embedding utilized. 
    Points plotted in the same color indicate the same instance across the dataset. Best viewed in color.
    }
    \label{fig:tsne}
  \end{minipage}
  \hspace{6mm}
  \begin{minipage}{.5\linewidth}
   \centering
\includegraphics[width=1.0\linewidth]{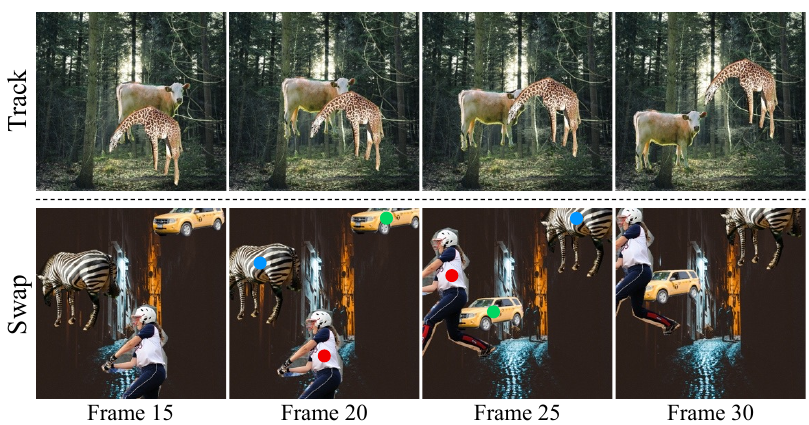}
\caption{
\textbf{Visualization of the pseudo dataset.}
In \textit{track type} videos, instances move along random bezier curves.
On the other hand, the \textit{swap type} refers to a scenario where the positions of each instance are exchanged in the middle of the video. The colored dot above each instance represents the corresponding instance in the swapped frame.
}
\label{fig:pseudo}
  \end{minipage}
  \vspace{-5mm}
\end{figure}

%% file: tab/pseudo.tex

\begin{table}[t]
    \centering
    \caption{Comparisons on \textbf{Pseudo dataset}.}
    \vspace{-2mm}
    \begin{tabularx}{0.8\textwidth}{Y|XX|Y|YYYYY}
        \toprule
        \multicolumn{3}{l|}{Method}   & App           & AP        & AP$_{50}$     & AP$_{75}$     & AR$_{1}$      & AR$_{10}$ \\
        \midrule
        \midrule
        
        \multirow{4}{*}{\rotatebox{90}{Track}}
        & \multicolumn{2}{l|}{GenVIS~\cite{GenVIS}}   &                & 54.8      & 70.6      & 58.8      & 61.2      & 65.9   \\ 
        & \multicolumn{2}{l|}{CTVIS~\cite{CTVIS}}        &                & 63.7      & 78.7      & 69.7      & 70.3      & 74.8   \\ 
        & \multicolumn{2}{l|}{\name{}}                   &                & 65.2      & 80.9      & 71.3      & 70.2      & 75.2   \\ 
        & \multicolumn{2}{l|}{\name{}}                   & \cmark         & \textbf{65.8}      & \textbf{81.1}          & \textbf{71.7}          & \textbf{70.8}          & \textbf{76.3}   \\ 
        \midrule
        \multirow{4}{*}{\rotatebox{90}{Swap}}
        & \multicolumn{2}{l|}{GenVIS~\cite{GenVIS}}      &                & 41.8      & 59.7      & 44.0      & 50.8      & 57.0   \\ 
        & \multicolumn{2}{l|}{CTVIS~\cite{CTVIS}}        &                & 53.1      & 71.7      & 56.9      & 61.7      & 65.7   \\ 
        & \multicolumn{2}{l|}{\name{}}                   &                & 51.8      & 72.1      & 54.4      & 57.7      & 62.8   \\ 
        & \multicolumn{2}{l|}{\name{}}                   & \cmark         & \textbf{63.0}      & \textbf{79.5}          & \textbf{69.3}          & \textbf{67.8}          & \textbf{73.5}   \\ 
        \bottomrule
    \end{tabularx}
    \vspace{-3mm}
    \label{tab:pseudo}
\end{table}

%% file: fig/qualitative.tex
\begin{figure*}[t]
\begin{center}
\includegraphics[width=\linewidth, bb=0 0 850 388]{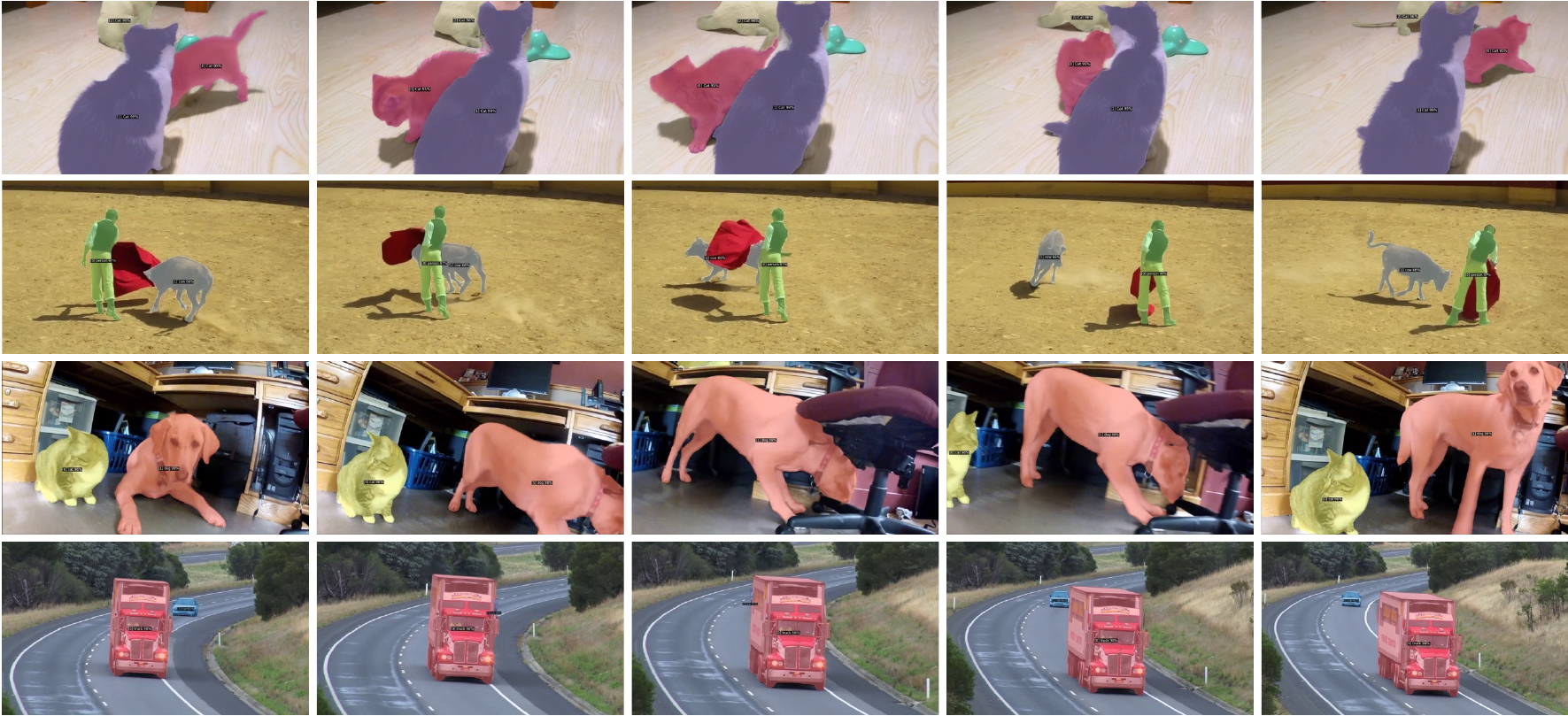}
\end{center}
\vspace{-6mm}
\caption{
\textbf{Qualitative results of VISAGE.} 
Videos are sourced from OVIS~\cite{OVIS-Dataset} and YTVIS 2021~\cite{vis2021} datasets. 
These videos represent complex scenarios, characterized by intersections and reappearances. Best viewed in color.
}
\vspace{-4mm}
\label{fig:qualitative}
\end{figure*}

%% file: sec/5_limitations.tex
\section{Limitations and Future Works}
\name{} addresses the lack of appearance awareness in existing query-based online VIS methods. 
Nevertheless, our approach encounters two main limitations. Firstly, while \name{} primarily employs a query-matching strategy, we recognize that query-propagation methods similarly depend on location information. 
This observation points towards a future research direction: integrating appearance information awareness into query-propagation approaches. 
Secondly, a fundamental issue not unique to our method but prevalent in tracking-by-detection approaches, especially in recent query-based VIS methods, is the heavy reliance on frame-level detectors.
During the tracking stage, these methods do not account for their own errors, allowing any inaccuracies to adversely affect video-level predictions. 
Addressing this challenge requires thorough consideration and represents another avenue for future work.

%% file: sec/6_conclusion.tex
\section{Conclusion}
In this paper, we explore the importance of appearance in tracking objects, an aspect often taken for granted yet overlooked by current online VIS methods.
We simply leverage the appearance cue when tracking objects, enabling more effective distinction of objects in challenging scenarios.
Our method, while simple yet effective, achieves performance comparable to previous methods across various VIS benchmarks.
However, as existing VIS benchmarks do not focus on appearance-requiring scenarios, we generate a synthetic dataset that necessitates the use of appearance information.
This dataset serves to validate our appearance-aware approach and our approach surpasses other methods with very large margin.
We believe that recognizing and leveraging the importance of appearance can lead to progress of VIS.

%% file: main.bbl
\begin{thebibliography}{10}
\providecommand{\url}[1]{\texttt{#1}}
\providecommand{\urlprefix}{URL }
\providecommand{\doi}[1]{https://doi.org/#1}

\bibitem{StEm-Seg}
Athar, A., Mahadevan, S., O{\v{s}}ep, A., Leal-Taix{\'e}, L., Leibe, B.: Stem-seg: Spatio-temporal embeddings for instance segmentation in videos. In: ECCV (2020)

\bibitem{DETR}
Carion, N., Massa, F., Synnaeve, G., Usunier, N., Kirillov, A., Zagoruyko, S.: End-to-end object detection with transformers. In: ECCV (2020)

\bibitem{SimCLR}
Chen, T., Kornblith, S., Norouzi, M., Hinton, G.: A simple framework for contrastive learning of visual representations. In: ICML (2020)

\bibitem{Mask2Former-VIS}
Cheng, B., Choudhuri, A., Misra, I., Kirillov, A., Girdhar, R., Schwing, A.G.: Mask2former for video instance segmentation. arXiv preprint arXiv:2112.10764  (2021)

\bibitem{Mask2Former}
Cheng, B., Misra, I., Schwing, A.G., Kirillov, A., Girdhar, R.: Masked-attention mask transformer for universal image segmentation. In: CVPR (2022)

\bibitem{CAROQ}
Choudhuri, A., Chowdhary, G., Schwing, A.G.: Context-aware relative object queries to unify video instance and panoptic segmentation. In: CVPR (2023)

\bibitem{ViT}
Dosovitskiy, A., Beyer, L., Kolesnikov, A., Weissenborn, D., Zhai, X., Unterthiner, T., Dehghani, M., Minderer, M., Heigold, G., Gelly, S., Uszkoreit, J., Houlsby, N.: An image is worth 16x16 words: Transformers for image recognition at scale. In: ICLR (2021)

\bibitem{QDTrack_TPAMI}
Fischer, T., Huang, T., Pang, J., Qiu, L., Chen, H., Darrell, T., Yu, F.: Qdtrack: Quasi-dense similarity learning for appearance-only multiple object tracking. TPAMI  (2023)

\bibitem{CopyPaste}
Ghiasi, G., Cui, Y., Srinivas, A., Qian, R., Lin, T.Y., Cubuk, E.D., Le, Q.V., Zoph, B.: Simple copy-paste is a strong data augmentation method for instance segmentation. In: CVPR (2021)

\bibitem{FastRCNN}
Girshick, R.: Fast r-cnn. In: ICCV (2015)

\bibitem{VISOLO}
Han, S.H., Hwang, S., Oh, S.W., Park, Y., Kim, H., Kim, M.J., Kim, S.J.: Visolo: Grid-based space-time aggregation for efficient online video instance segmentation. In: CVPR (2022)

\bibitem{InsPro}
He, F., Zhang, H., Gao, N., Jia, J., Shan, Y., Zhao, X., Huang, K.: Inspro: Propagating instance query and proposal for online video instance segmentation. In: NeurIPS (2022)

\bibitem{MaskRCNN}
He, K., Gkioxari, G., Dollar, P., Girshick, R.: Mask r-cnn. In: ICCV (2017)

\bibitem{ResNet}
He, K., Zhang, X., Ren, S., Sun, J.: Deep residual learning for image recognition. In: CVPR (2016)

\bibitem{GenVIS}
Heo, M., Hwang, S., Hyun, J., Kim, H., Oh, S.W., Lee, J.Y., Kim, S.J.: A generalized framework for video instance segmentation. In: CVPR (2023)

\bibitem{VITA}
Heo, M., Hwang, S., Oh, S.W., Lee, J.Y., Kim, S.J.: Vita: Video instance segmentation via object token association. In: NeurIPS (2022)

\bibitem{MinVIS}
Huang, D.A., Yu, Z., Anandkumar, A.: Minvis: A minimal video instance segmentation framework without video-based training. In: NeurIPS (2022)

\bibitem{IFC}
Hwang, S., Heo, M., Oh, S.W., Kim, S.J.: Video instance segmentation using inter-frame communication transformers. In: NeurIPS (2021)

\bibitem{PCAN}
Ke, L., Li, X., Danelljan, M., Tai, Y.W., Tang, C.K., Yu, F.: Prototypical cross-attention networks for multiple object tracking and segmentation. In: NeurIPS (2021)

\bibitem{VPS}
Kim, D., Woo, S., Lee, J.Y., Kweon, I.S.: Video panoptic segmentation. In: CVPR (2020)

\bibitem{Hungarian}
Kuhn, H.W.: The hungarian method for the assignment problem. NRL  (1955)

\bibitem{BG-20K}
Li, J., Zhang, J., Maybank, S.J., Tao, D.: Bridging composite and real: Towards end-to-end deep image matting. IJCV  (2022)

\bibitem{TCOVIS}
Li, J., Yu, B., Rao, Y., Zhou, J., Lu, J.: Tcovis: Temporally consistent online video instance segmentation. In: ICCV (2023)

\bibitem{COCO}
Lin, T.Y., Maire, M., Belongie, S., Hays, J., Perona, P., Ramanan, D., Doll{\'a}r, P., Zitnick, C.L.: Microsoft coco: Common objects in context. In: ECCV (2014)

\bibitem{OVIS}
Qi, J., Gao, Y., Hu, Y., Wang, X., Liu, X., Bai, X., Belongie, S., Yuille, A., Torr, P.H., Bai, S.: Occluded video instance segmentation. arXiv preprint arXiv:2102.01558  (2021)

\bibitem{OVIS-Dataset}
Qi, J., Gao, Y., Hu, Y., Wang, X., Liu, X., Bai, X., Belongie, S., Yuille, A., Torr, P.H., Bai, S.: Occluded video instance segmentation: A benchmark. IJCV  (2022)

\bibitem{ContrastiveNIPS}
Sohn, K.: Improved deep metric learning with multi-class n-pair loss objective. In: NeurIPS (2016)

\bibitem{VisTR}
Wang, Y., Xu, Z., Wang, X., Shen, C., Cheng, B., Shen, H., Xia, H.: End-to-end video instance segmentation with transformers. In: CVPR (2020)

\bibitem{EfficientVIS}
Wu, J., Yarram, S., Liang, H., Lan, T., Yuan, J., Eledath, J., Medioni, G.: Efficient video instance segmentation via tracklet query and proposal. In: CVPR (2022)

\bibitem{SeqFormer}
Wu, J., Jiang, Y., Zhang, W., Bai, X., Bai, S.: Seqformer: a frustratingly simple model for video instance segmentation. In: ECCV (2022)

\bibitem{IDOL}
Wu, J., Liu, Q., Jiang, Y., Bai, S., Yuille, A., Bai, X.: In defense of online models for video instance segmentation. In: ECCV (2022)

\bibitem{MaskTrackRCNN}
Yang, L., Fan, Y., Xu, N.: Video instance segmentation. In: ICCV (2019)

\bibitem{vis2021}
Yang, L., Fan, Y., Xu, N.: The 3rd large-scale video object segmentation challenge - video instance segmentation track (2021)

\bibitem{vis2022}
Yang, L., Fan, Y., Xu, N.: The 4th large-scale video object segmentation challenge - video instance segmentation track (2022)

\bibitem{CrossVIS}
Yang, S., Fang, Y., Wang, X., Li, Y., Fang, C., Shan, Y., Feng, B., Liu, W.: Crossover learning for fast online video instance segmentation. In: ICCV (2021)

\bibitem{TeViT}
Yang, S., Wang, X., Li, Y., Fang, Y., Fang, J., Liu, W., Zhao, X., Shan, Y.: Temporally efficient vision transformer for video instance segmentation. In: CVPR (2022)

\bibitem{CTVIS}
Ying, K., Zhong, Q., Mao, W., Wang, Z., Chen, H., Wu, L.Y., Liu, Y., Fan, C., Zhuge, Y., Shen, C.: Ctvis: Consistent training for online video instance segmentation. In: ICCV (2023)

\bibitem{DVIS}
Zhang, T., Tian, X., Wu, Y., Ji, S., Wang, X., Zhang, Y., Wan, P.: Dvis: Decoupled video instance segmentation framework. In: ICCV (2023)

\bibitem{Deformable-DETR}
Zhu, X., Su, W., Lu, L., Li, B., Wang, X., Dai, J.: Deformable detr: Deformable transformers for end-to-end object detection. In: ICLR (2021)

\end{thebibliography}
